%% file: acl24-appropriateness-style-transfer-frame.tex
\pdfoutput=1

\documentclass[11pt]{article}

\usepackage[]{acl}

\usepackage{times}
\usepackage{latexsym}

\usepackage[T1]{fontenc}

\usepackage[utf8]{inputenc}

\usepackage{microtype}

\usepackage{inconsolata}

\usepackage{lipsum}  
\usepackage{tabularx}
\usepackage{booktabs}

\usepackage{microtype}
\usepackage{type1cm}
\usepackage[american]{babel}
\usepackage{amssymb}
\usepackage{amsmath}
\usepackage{multirow}
\usepackage{xspace}
\DeclareMathAlphabet{\mathcalbf}{OMS}{pzc}{b}{n}
\usepackage{enumitem}
\usepackage{colortbl}

\RequirePackage{color}
\definecolor{darkgray}{gray}{0.40}
\definecolor{mediumgray}{gray}{0.60}
\definecolor{lightgray}{gray}{0.95}
\definecolor{ultralightgray}{gray}{0.98}
\definecolor{forestgreen}{rgb}{0.133, 0.545, 0.133}
\definecolor{orange}{rgb}{1, 0.86, 0.74}
\definecolor{lightergreen}{rgb}{0.95, 1, 0.88}

\usepackage{graphicx}
\DeclareGraphicsExtensions{.pdf,.ai,.jpg,.png}
\setkeys{Gin}{pagebox=artbox}
\graphicspath{{../acl24-appropriateness-style-transfer-figures/}}

\newcommand{\bsfigure}[3][]{%
    \begin{figure}[t]
        \centering
        \includegraphics[#1]{#2}
        \caption{#3}\label{#2}%
    \end{figure}
}

\RequirePackage{type1cm}
\RequirePackage{color}
\RequirePackage{soul}
\setstcolor{blue}
\definecolor{violet}{rgb}{0.5,0.0,0.5}

\newsavebox\bscombox
\newcommand{\bscom}[3][]{%
    \sbox{\bscombox}{\fontsize{8}{9}\selectfont#1#2#3}
    \noindent
    \st{#2}{\selectfont
        \color{blue}#3\ifx\\#1\\\else{\fontsize{8}{9}\selectfont\color{violet}[#1]}\fi
    }
}

\RequirePackage[normalem]{ulem} 
\definecolor{RED}{RGB}{216,27,97}\definecolor{BLUE}{RGB}{0,157,115} 
\providecommand{\DIFaddtex}[1]{{\protect\color{BLUE}#1}} 
\providecommand{\DIFdeltex}[1]{{\protect\color{RED}\sout{#1}}}                      
\providecommand{\DIFaddbegin}{} 
\providecommand{\DIFaddend}{} 
\providecommand{\DIFdelbegin}{} 
\providecommand{\DIFdelend}{} 
\providecommand{\DIFaddbeginFL}{} 
\providecommand{\DIFaddendFL}{} 
\providecommand{\DIFdelbeginFL}{} 
\providecommand{\DIFdelendFL}{} 
\providecommand{\DIFadd}[1]{\texorpdfstring{\DIFaddtex{#1}}{#1}} 
\providecommand{\DIFdel}[1]{\texorpdfstring{\DIFdeltex{#1}}{}} 
\newcommand{\DIFscaledelfig}{0.5}
\RequirePackage{settobox} 
\RequirePackage{letltxmacro} 
\newsavebox{\DIFdelgraphicsbox} 
\newlength{\DIFdelgraphicswidth} 
\newlength{\DIFdelgraphicsheight} 
\LetLtxMacro{\DIFOincludegraphics}{\includegraphics} 
\newcommand{\DIFaddincludegraphics}[2][]{{\color{BLUE}\fbox{\DIFOincludegraphics[#1]{#2}}}} 
\newcommand{\DIFdelincludegraphics}[2][]{
\sbox{\DIFdelgraphicsbox}{\DIFOincludegraphics[#1]{#2}}
\settoboxwidth{\DIFdelgraphicswidth}{\DIFdelgraphicsbox} 
\settoboxtotalheight{\DIFdelgraphicsheight}{\DIFdelgraphicsbox} 
\scalebox{\DIFscaledelfig}{
\parbox[b]{\DIFdelgraphicswidth}{\usebox{\DIFdelgraphicsbox}\\[-\baselineskip] \rule{\DIFdelgraphicswidth}{0em}}\llap{\resizebox{\DIFdelgraphicswidth}{\DIFdelgraphicsheight}{
\setlength{\unitlength}{\DIFdelgraphicswidth}
\begin{picture}(1,1)
\thicklines\linethickness{2pt} 
{\color[rgb]{1,0,0}\put(0,0){\framebox(1,1){}}}
{\color[rgb]{1,0,0}\put(0,0){\line( 1,1){1}}}
{\color[rgb]{1,0,0}\put(0,1){\line(1,-1){1}}}
\end{picture}
}\hspace*{3pt}}} 
} 
\LetLtxMacro{\DIFOaddbegin}{\DIFaddbegin} 
\LetLtxMacro{\DIFOaddend}{\DIFaddend} 
\LetLtxMacro{\DIFOdelbegin}{\DIFdelbegin} 
\LetLtxMacro{\DIFOdelend}{\DIFdelend} 
\DeclareRobustCommand{\DIFaddbegin}{\DIFOaddbegin \let\includegraphics\DIFaddincludegraphics} 
\DeclareRobustCommand{\DIFaddend}{\DIFOaddend \let\includegraphics\DIFOincludegraphics} 
\DeclareRobustCommand{\DIFdelbegin}{\DIFOdelbegin \let\includegraphics\DIFdelincludegraphics} 
\DeclareRobustCommand{\DIFdelend}{\DIFOaddend \let\includegraphics\DIFOincludegraphics} 
\LetLtxMacro{\DIFOaddbeginFL}{\DIFaddbeginFL} 
\LetLtxMacro{\DIFOaddendFL}{\DIFaddendFL} 
\LetLtxMacro{\DIFOdelbeginFL}{\DIFdelbeginFL} 
\LetLtxMacro{\DIFOdelendFL}{\DIFdelendFL} 
\DeclareRobustCommand{\DIFaddbeginFL}{\DIFOaddbeginFL \let\includegraphics\DIFaddincludegraphics} 
\DeclareRobustCommand{\DIFaddendFL}{\DIFOaddendFL \let\includegraphics\DIFOincludegraphics} 
\DeclareRobustCommand{\DIFdelbeginFL}{\DIFOdelbeginFL \let\includegraphics\DIFdelincludegraphics} 
\DeclareRobustCommand{\DIFdelendFL}{\DIFOaddendFL \let\includegraphics\DIFOincludegraphics} 
\RequirePackage{listings} 
\RequirePackage{color} 
\lstdefinelanguage{DIFcode}{ 
  moredelim=[il][\color{RED}\sout]{\%DIF\ <\ }, 
  moredelim=[il][\color{BLUE}\uwave]{\%DIF\ >\ } 
} 
\lstdefinestyle{DIFverbatimstyle}{ 
	language=DIFcode, 
	basicstyle=\ttfamily, 
	columns=fullflexible, 
	keepspaces=false 
} 
\lstnewenvironment{DIFverbatim}{\lstset{style=DIFverbatimstyle}}{} 
\lstnewenvironment{DIFverbatim*}{\lstset{style=DIFverbatimstyle,showspaces=true}}{} 
\begin{document}

\input{acl24-appropriateness-style-transfer-pre}
\input{acl24-appropriateness-style-transfer-part1}

\input{acl24-appropriateness-style-transfer-part2}

\input{acl24-appropriateness-style-transfer-part3}

\input{acl24-appropriateness-style-transfer-part4}
\input{acl24-appropriateness-style-transfer-part5}
\input{acl24-appropriateness-style-transfer-sum}
\input{acl24-appropriateness-style-transfer-ack}
\input{acl24-appropriateness-style-transfer-limitations}
\input{acl24-appropriateness-style-transfer-ethical}

\bibliography{acl24-appropriateness-style-transfer-lit}
\appendix
\input{appendix}

\end{document}

%% file: acl24-appropriateness-style-transfer-pre.tex

\title{LLM-based Rewriting of Inappropriate Argumentation \\ using Reinforcement Learning from Machine Feedback}

\author{
    Timon Ziegenbein \\
    Leibniz University Hannover \\
    \texttt{t.ziegenbein@ai.uni-hannover.de} \\\And
    Gabriella Skitalinskaya \\
    Leibniz University Hannover \\
    \texttt{g.skitalinska@ai.uni-hannover.de} \\\AND
    Alireza Bayat Makou \\
    Leibniz University Hannover \\
    \texttt{a.bayat.makou@stud.uni-hannover.de} \\\And
    Henning Wachsmuth \\
    Leibniz University Hannover \\
    \texttt{h.wachsmuth@ai.uni-hannover.de} \\\
}

\maketitle

\begin{abstract}
Ensuring that online discussions are civil and productive is a major challenge for social media platforms. Such platforms usually rely both on users and on automated detection tools to flag inappropriate arguments of other users, which moderators then review. However, this kind of post-hoc moderation is expensive and time-consuming, and moderators are often overwhelmed by the amount and severity of flagged content. Instead, a promising alternative is to prevent negative behavior during content creation. This paper studies how inappropriate language in arguments can be computationally mitigated. We propose a reinforcement learning-based rewriting approach that balances content preservation and appropriateness based on existing classifiers, prompting an instruction-finetuned large language model (LLM) as our initial policy. Unlike related style transfer tasks, rewriting inappropriate arguments allows deleting and adding content permanently. It is therefore tackled on document level rather than sentence level. We evaluate different weighting schemes for the reward function in both absolute and relative human assessment studies. Systematic experiments on non-parallel data provide evidence that our approach can mitigate the inappropriateness of arguments while largely preserving their content. It significantly outperforms competitive baselines, including few-shot learning, prompting, and humans.
\end{abstract}

%% file: acl24-appropriateness-style-transfer-part1.tex
\section{Introduction}
\label{sec:introduction}

\bsfigure{rewrite-example}{Example of an inappropriate argument from the corpus of \citet{ziegenbein:2023} and the same argument after applying our approach. The used colors indicate which parts of the original argument were removed (red strikethrough) and which parts were added by our approach in the rewriting process (green).}

Creating trusted and safe online spaces where people with different backgrounds and opinions can discuss controversial issues is a major challenge for social media platforms \cite{salminen:2018}. The diversity in opinions, emotional attachments, and the anonymity of the web easily lead to heated discussions, which can quickly turn into toxic environments, even if only one participant behaves \emph{inappropriately} \cite{habernal:2018}. Avoiding this is a challenging task, often supported by platform moderators that check content flagged by users or detection tools. 
However, the amount of moderation required on the web necessitates automation of the process, as the resources for manual moderation are usually insufficient, and the severity of inappropriate content can negatively affect the moderators' psyche \cite{spence:2023}.

Multiple concepts and datasets have been proposed to model unwanted behavior in discussions, from simple offensiveness \cite{borkan:2019} to  sophisticated notions such as inappropriateness \cite{wachsmuth:2017a}. The latter focuses on the exchange of arguments, their creation of credibility and emotions, and their adherence to the issue. \citet{ziegenbein:2023} argue that appropriateness displays the minimal quality of an argument necessary to be considered valuable in a debate.

This paper is the first to study how to rewrite inappropriate arguments automatically. Prior work studied the detection of unwanted behavior \cite{wulczyn:2016,he:2023,ziegenbein:2023}, often using large language models (LLM). While a few methods improve content, they solely transfer the style of texts to be more formal \cite{rao:2018,lai:2021a}, less subjective \cite{pryzant:2020,liu:2021a}, or less toxic \cite{laugier:2021,logacheva:2022}, or they target the quality of arguments in general \cite{skitalinskaya:2023}. This commonly comes with preserving the original content and operating on single sentences. However, if the inappropriate behavior is rooted in the content itself and not only in the style of the text, content modifications on the document level may be necessary. In addition, most existing approaches rely on parallel data, which is unavailable for rewriting inappropriate arguments.\,\,

Instead, we propose an LLM-based rewriting approach to inappropriateness mitigation inspired by reinforcement learning from human feedback, RLHF \cite{christiano:2017}. Compared to the typical use of RLHF in NLP \cite{ouyang:2022}, the core ideas of our approach are: (1) We obtain a `cheap' initial policy (an LLM to align) from either few-shot learning or prompting, rather than using supervised learning. (2) We specifically consider the properties on which we want to align the LLM, rather than relying on generic preference information. (3) We evaluate multiple candidate alignments with different weightings of the desired properties, rather than relying~on a single alignment obtained from preference information.

After experimentally determining the performance of multiple LLMs using few-shot learning and prompting on the corpus of \citet{ziegenbein:2023}, we find prompting the instruction-finetuned LLaMA~\cite{touvron:2023} variant of \citet{taori:2023} (Alpaca) to be best and thus proceed to align it further using our approach.
We deem the desired properties for rewriting inappropriate arguments to be \emph{semantic similarity} to the original argument and \emph{appropriateness} of the generated argument, and we make use of existing classifiers to learn how to generate texts that fulfill them \cite{zhang:2020,ziegenbein:2023}.

Exemplarily, Figure \ref{rewrite-example} shows an inappropriate argument from a ``Pro choice vs pro life'' debate and the same argument rewritten by our approach. Here, the original argument uses overly excessive emotions, making it hard to understand, and it displays little interest in the opinion of others. The rewritten argument reduces emotions and adds a more open ending, making it more appropriate while keeping the original argument's gist intact. 

For evaluation, we obtain human rewrites for a portion of the data and compare them automatically against our models and a competitive model from the literature. Moreover, we conduct relative and absolute human evaluations of the rewrites of our trained models and the human rewrites. We find that our approach successfully aligns LLMs according to the desired property weighting and produces the best rewrites. Intriguingly, our human annotators prefer appropriate rewrites, even if they are less semantically similar to the original arguments.


Altogether, this paper's main contributions are:%
\footnote{The corpus extension and experiment code can be found under: \url{https://github.com/webis-de/ACL-24}.}
\begin{itemize}
\setlength{\itemsep}{0ex}
\item
An RLHF-inspired approach for non-parallel data, based on instruction-finetuned LLMs aligned to specific classified properties.
\item
The first computational approach for rewriting inappropriate arguments.
\item
Empirical insights into human preferences regarding semantic similarity and appropriateness when rewriting inappropriate arguments.
\end{itemize}

%% file: acl24-appropriateness-style-transfer-part2.tex
\section{Related Work}
\label{sec:related-work}

The notion of appropriateness in speech and argumentation, tied to cultural norms, social politeness, and context, originates from Aristotle's work on rhetoric \cite{aristotle:2007}. It has been examined in various shades across linguistic studies \cite{hymes:1972, ranney:1992, schneider:2012, jdetawy:2020}. In debate, topic adherence and avoidance of offensive or biased language are considered aspects of appropriateness \cite{andrew:1996, blair:1999, walton:1999, burkett:2011}. Modeled by \citet{wachsmuth:2017a} as a dimension of rhetorical argument quality, appropriateness has been partially explored in NLP, focusing on the simultaneous assessment of credibility, emotional engagement, and proportionality to the issue. Computationally, \citet{wachsmuth:2020} initially attempted appropriateness prediction as a subtask of argument quality assessment. Later, \citet{ziegenbein:2023} refined the notion of appropriateness in argumentation, modeling it in a 14-dimensional taxonomy and predicting it together with its subdimensions in a multilabel setting. \citet{wachsmuth:2024} recently delineated how to instruct LLMs towards more reliable argument quality assessment, not targeting rewriting though.

Related to argument rewriting, \citet{skitalinskaya:2023} studied how to improve the general quality of argumentative claims on data similar to what we use, but not focusing on inappropriateness specifically. Closest to our work in the context of style transfer tasks are \citet{nogueira:2018}, \citet{laugier:2021}, and \citet{he:2023}, where toxic content is mitigated using supervised rewriting approaches. 
However, their methods are applied on the sentence level, strictly aim to preserve content, and prevent the addition of new content. Unlike them, we focus on document-level rewriting and explicitly consider adding or deleting content. Furthermore, most of these approaches require parallel data and rely on supervised learning. In contrast, our approach is meant for non-parallel data since no parallel dataset is available to learn to rewrite inappropriate arguments.

To this end, we use reinforcement learning as it allows us to train on non-differentiable metrics, such as the outputs of classifiers, which we use to test for the desired properties of our task. Reinforcement learning has been used for a variety of NLP tasks, including dialogue generation \cite{li:2016}, machine translation \cite{wu:2018}, and summarization \cite{ziegler:2019,bohm:2019,stiennon:2020}. In the context of style transfer, multiple approaches have aimed to flip the sentiment, stance, or polarity of texts in parallel \cite{abhilasha:2020,liu:2021b} and non-parallel settings \cite{xu:2018,gong:2019,wu:2019,luo:2019}. 

Many works study the related task of formality transfer \cite{gong:2019,luo:2019,abhilasha:2020,liu:2021b,lai:2021a} with diverse techniques on the parallel data of \newcite{rao:2018}. Three properties are commonly controlled during transfer: fluency, content preservation, and transfer strength. Similar to our work, \citet{madanagopalr:2023} propose reinforcement learning to remove subjective bias in Wikipedia texts, modeling the reward as a weighted function of classifiers for style, fluency, and content preservation. However, unlike in formality transfer, parallel data for rewriting inappropriate arguments is neither available nor straightforward to acquire. 

Consequently, our work focuses on non-parallel style transfer inspired by reinforcement learning from human feedback. We rely on the more stable proximal policy optimization (PPO) \cite{schulman:2017} instead of the commonly used REINFORCE Monte Carlo policy gradient \cite{williams:1992} due to promising recent results in NLP and the advances in the capabilities of language models to follow instructions \cite{taori:2023,wang:2023}. To our knowledge, only the work of \citet{de:2024}, which was published on arXiv after the submission of our work, uses an RLHF-inspired PPO approach for a style transfer-related task in NLP. Our work is the first to investigate the use of prompting as an initial policy, avoiding the need for parallel data through pseudo-parallel data and applying RLHF-inspired PPO to non-parallel data.

\subsection{Proximal Policy Optimization in NLP}
\label{sec:ppo}

As a basis for the presentation of our approach, we here shortly describe the intuition behind PPO in an NLP context. For a more formal description, we refer the reader to \citet{zheng:2023}.

PPO learns a \emph{model critique} (a value model), which estimates the expected cumulative future \emph{reward} (a value) of a \emph{state} (the generated text up to this point), together with a \emph{policy} (e.g., an LLM). The ``real'' reward is based on the output of a \emph{reward model} (e.g., a classifier) and can be any scalar value. The value model estimates the \emph{advantage} (gain in reward) of performing a specific \emph{action} in a state (generating a specific word given the text generated so far) over performing the current policy's suggested action. Specific actions are chosen based on sampling from the current policy (e.g., top-p sampling). The current policy is updated based on the advantages of the specific actions and the KL-divergence between the token-level distributions of the current policy and its updated version. The KL-divergence here improves stability during training, limiting the size of update steps.

In other words, when predicting the next word given a sequence of previously generated words, multiple fitting potential words are considered by sampling from the LLM. The difference in the long-term expected reward between each of them and the word with the current highest probability is used to update the LLM. This way, the LLM can be steered (aligned) to generate text that fulfills the desired properties represented by the reward model.

%% file: acl24-appropriateness-style-transfer-part3.tex
\section{Approach}  
\label{sec:approach}

This section presents the approach that we propose to rewrite inappropriate arguments inspired by reinforcement learning from human feedback. Figure~\ref{rewrite-approach} illustrates the main elements of our approach explained in the following.

\subsection{Problem Formulation}
\label{sec:problem-formulation}

Let $x$ be an argument and $\hat{y}$ be an improved version of $x$. We define the task of \emph{rewriting inappropriate arguments} as learning a function $f: x \mapsto \hat{y}$ such that $\hat{y}$ preserves the content of $x$ as much as possible while being more appropriate. 

\bsfigure{rewrite-approach}{Our approach to rewriting inappropriate arguments: The policy $\pi^{RL}$ is optimized using \emph{PPO} to generate an improved version $\hat{y}$ from the input argument $x$ while preserving the content of $x$ as much as possible ($c_{sim}$) and making the argument more appropriate ($c_{app}$). This is based on reward $R$ obtained from the weighting of $r$ of the scalar classifier outputs and the \emph{KL-divergence} between the initial policy $\pi^{PRT}$ and the current $\pi^{RL}$. Dashed lines: The probability distribution over the tokens is used as the output of the LLM.}

\subsection{Prompting as an Initial Policy}
\label{sec:obtain-initial-policy}

Usually, reinforcement learning from human feedback (RLHF) is used only to steer an LLM that has already learned to solve a task to a certain known extent instead of learning the task from scratch. However, we neither have access to such a model, nor data to train it. Instead, we thus propose to use prompting to obtain an initial policy $f~\hat{=}~\pi^{PRT}$ (an LLM to start from) that solves the task to a certain (not immediately quantifiable) extent. We experimentally compare multiple autoregressive pretrained LLMs using zero-shot, few-shot, and instruction learning-based prompting to find an effective initial policy (further details on the models and prompting methods follow in Section~\ref{sec:experiments}). Then, we proceed to use the same prompts while learning $\pi_{\phi}^{RL}$, where $\phi$ are the learnable parameters, to obtain a better version of the initial policy $\pi^{PRT}$.%
\footnote{In the following, we use the terms \emph{policy} and \emph{LLM} interchangeably as they refer to the same concept ($\pi$) in the context of reinforcement learning (RL) for LLMs.}

\subsection{Reward Modeling and Policy Learning}
\label{sec:policy-learning}

We initialize $\pi_{\phi}^{RL}$ with a pretrained LLM $\pi^{PRT}$ that is prompted in natural language to generate~$\hat{y}$ given~$x$. Unlike \citet{stiennon:2020}, we do not require learning a reward model $r$ from human feedback on preference judgments. Instead, we model relevant properties that we desire to be present in the target output, $\hat{y}$. In particular, we assume that the semantic similarity ($sim$) of $\hat{y}$ to $x$ and the appropriateness ($app$) of $\hat{y}$ are such relevant properties. We use pretrained classifiers $c$ that estimate these properties as our reward model. 

The property reward model $r$ is thus defined as:
\begin{equation}
    r(x,\hat{y}) := \alpha \cdot {c_{sim}}(x,\hat{y}) + (1-\alpha) \cdot {c_{app}}(\hat{y})
\end{equation}

Here, $\alpha \in [0, 1]$ is a hyperparameter that controls the trade-off between semantic similarity and appropriateness. $c_{sim}$ and $c_{app}$ are classifiers that estimate the semantic similarity and appropriateness of $\hat{y}$ given $x$.
Similar to \citet{stiennon:2020}, to obtain the final reward model $R$, we penalize $r$ with the KL-divergence between the initial policy $\pi^{PRT}$ and the learned policy $\pi_{\phi}^{RL}$ to disincentivize moving away too far from $\pi^{PRT}$:
\begin{equation}
    R(x,\hat{y}) := r(x,\hat{y}) - \beta \log \left[\frac{\pi_{\phi}^{RL}(\hat{y} \vert x)}{\pi^{PRT}(\hat{y} \vert x)}\right]
\end{equation}

\noindent
Here, $\beta \in \mathbb{R}$ is a hyperparameter that controls the strength of the KL-divergence. The policy $\pi_{\phi}^{RL}$ is optimized using the proximal policy optimization (PPO) algorithm \cite{schulman:2017}.

We argue our setup to be beneficial for two reasons: (1)~Pretrained classifiers that learned to assess specific properties from human labels are available for a wide variety of tasks; and (2)~no training and consequently no data is required to learn an initial policy that solves the task to a certain extent.

%% file: acl24-appropriateness-style-transfer-part4.tex
\section{Data}
\label{sec:data}

For our mitigation experiments, we extended the appropriateness corpus of \citet{ziegenbein:2023}.

\subsection{Source Data}

The original corpus contains 2191~arguments and the corresponding discussion titles from three genres (reviews, discussion forums, and Q\&A~forums). Each argument has been annotated three times using a hierarchical 14-dimensional taxonomy of appropriateness flaws, such as \emph{toxic emotions} or \emph{missing intelligibility}. Here, we consider only the parent dimension \emph{inappropriateness} to develop and evaluate our approaches. The corpus contains 1182 inappropriate and 1009 appropriate arguments.

\subsection{Extension}

We extend the given corpus by arguments from its original domains. In particular, we collected 73,703 arguments from the IACv2 Corpus \cite{walker:2012,abbott:2016} and the GAQCorpus \cite{ng:2020}. We kept only those 55,290 arguments that have at least 10 and at most 220 words and that do not exceed 1100 characters. This way, we ensure that the arguments have approximately the same length as those in the original corpus. To avoid any topic leakage in the extended part of the corpus, we also remove arguments that belong to a topic already present in the original version of the corpus (49,417 arguments remaining). Finally, we soft-label all arguments in the extended part of the corpus with the five-fold ensemble classifier of \citet{ziegenbein:2023} to obtain appropriateness labels. These labels can then be used to train our approaches. 35,537 of the arguments were labeled as inappropriate and 13,880 as appropriate.

%% file: acl24-appropriateness-style-transfer-part5.tex
\section{Experiments}    
\label{sec:experiments} 

This section describes the training procedure of our approach from Section~\ref{sec:approach} on the data fom Section~\ref{sec:data} and the experiments we conducted to evaluate it.

\subsection{Experimental Setup}

As \citet{ziegenbein:2023}, we split the data for evaluation into 70\% training, 10\% validation, and 20\% test, ensuring an equal weighting of the 14~corpus dimensions. However, since we are interested in mitigating inappropriateness, we train and evaluate only on inappropriate arguments. 

During training, we exclusively use the inappropriate arguments from the corpus extension. 

The advantage of this setup is two-fold: First, we can train on a large amount of data, which is often crucial for the success of reinforcement learning (RL). Second, it allows us to avoid propagating any selection bias that may arise from the intermediate step of finding the best initial policy to selecting the best checkpoint from our trained policy. We use the training set to select the best initial policy (details below), the validation set to select the best-performing RL policy checkpoints, and the test set to evaluate the performance of all approaches.

To automatically evaluate the generated rewrites, we employ the five appropriateness classifiers of \citet{ziegenbein:2023} trained on different folds of the data. As we use the entire appropriateness corpus for evaluation, we ensure the use of the classifier for each argument that has not seen the argument before as part of its training. We use a classifier to predict the appropriateness of the original argument and the generated rewrites. Then, we calculate the following performance values:
\begin{itemize}
\setlength{\itemsep}{-2pt}
\item
\emph{App.} Percentage of arguments for which an approach has flipped the prediction from inappropriate to appropriate;
\item
\emph{Sim.} Semantic similarity of a rewrite to the input argument in terms of BERTScore \cite{zhang:2020};
\item
\emph{NES.} Normalized word-wise edit similarity (to quantify amounts of edits) \cite{lopresti:1996};
\item
\emph{PPL.} Fluency in terms of perplexity;
\item
\emph{GM.} Geometric mean of \emph{App.}, \emph{Sim.} and \emph{1/PPL} (to compare approaches using a single score).
\end{itemize}

We use the semantic similarity and normalized word-wise edit similarity to quantify if the generated arguments are indeed rewrites of the original argument and not any probably unrelated but appropriate text. Furthermore, we use perplexity as a measure of text coherence and fluency.

\subsection{Finding an Initial Policy}
\label{sec:initial-policy}

Similar to \citet{stiennon:2020}, we start with obtaining the initial policy, $\pi^{PRT}$ that should have a reasonable performance mitigating inappropriate language. $\pi^{PRT}$ is then aligned using our RL approach.
Since no parallel data is available to train a supervised model, we prompt four LLMs of similar size (6-7 billion parameters) in a few-shot learning and instruction following-based setting:
\begin{itemize}
\setlength{\itemsep}{-4pt}
\item
\emph{OPT} \cite{zhang:2022};
\item
\emph{BLOOM} \cite{scao:2023};
\item
\emph{GPT-J.} \cite{wang:2021};
\item
\emph{LLaMA.} \cite{touvron:2023}.
\end{itemize}

For the few-shot setting, we use 1, 4, and 9 examples to see how the performance changes with the amount of reference data. This setup is inspired by the hierarchical setup of the taxonomy of inappropriateness \cite{ziegenbein:2023} having 1, 4, and 9 dimensions on the first, second, and third level respectively. To obtain instruction-finetuned versions of the models, we train each of them following the procedure suggested by \citet{taori:2023}, to ensure we select the best base model and do not select a model because of a specific way it is prompted or the instruction data it was fine-tuned on. This way, we have a single fixed prompt for all models and can control the generation length and other parameters equally well.

\paragraph{Creating Few-Shot Examples}

Since no rewrites are available for our few-shot learning setup, we collect 14 rewrites (1+4+9) of inappropriate arguments from three NLP experts, none of whom are authors of this paper. To create rewrites that are highly representative of a dimension, we use the appropriateness corpus, sentence transformers \cite{reimers:2019}, and PageRank \cite{lawrence:1998} (details in Appendix~\ref{sec:few-shot-examples}).

\paragraph{Prompting Setup}

We employ the natural language prompts suggested by \citet{reif:2022} and \citet{zhang:2020} to generate $\hat{y}$ given $x$. The full prompts can be found in Appendix~\ref{sec:prompting-setup}.

\input{table-initial-policy-automatic-evaluation.tex}

\paragraph{Automatic Evaluation}

Table~\ref{tab:initial-policy-automatic-evaluation} shows the results of finding an initial policy. We observe \emph{BLOOM + Instruct.}\ to create the most appropriate rewrites (0.653). \emph{LLaMA + Instruct.} achieves the best results in terms of semantic similarity (0.620), and zero-shot \emph{LLaMA} in terms of needing the minimal amount of edits to create its rewrites (0.528). In terms of fluency, \emph{GPT-J + Instruct.}\ performs best (37.49), closely followed by \emph{LLaMA + Instruct.}\ (38.08). Overall, \emph{LLaMA + Instruct.}\ seems to be the most stable choice (GM 0.216), so we select it as the initial policy to train our final approaches. Regarding geometric mean, few-shot learning leads to a general increase in performance. However, no clear trend in the number of few-shot examples used is visible. Overall, rewriting based on instruction-finetuning mitigates inappropriateness best across all models.

\subsection{Policy Learning}
\label{sec:rl-policy}

Starting from our initial policy, \emph{LLaMA + Instruct.}, we use PPO to learn a set of candidate policies. We use the rescaled version of BERTScore \cite{zhang:2020} to estimate the semantic similarity of $\hat{y}$ to $x$ and the appropriateness classifier from the first fold of \citet{ziegenbein:2023} to estimate the appropriateness of $\hat{y}$. Both values are in $[0, 1]$. We learn four candidate policies, each with a different property weighting $\alpha \in \{0.4, 0.5, 0.6, 1\}$. Indicating the property weighting used, we refer to the corresponding models as \emph{LLaMA + PPO$_{app<sim}$}, \emph{LLaMA + PPO$_{app=sim}$}, \emph{LLaMA + PPO$_{app>sim}$} and \emph{LLaMA + PPO$_{app}$} respectively.
The exact setup of our PPO training and the hyperparameters used are detailed in Appendix~\ref{sec:hyperparameters}.


\input{table-automatic-evaluation.tex}

\paragraph{Baselines}

In addition to the learned policies, we collected human rewrites for each argument. We refer to these as \emph{Human Baseline}. For this purpose, we hired five native English speakers on Upwork.com, three male and two female. We instructed the annotators with background information about appropriateness and asked them to suggest rewritten versions of arguments flagged as inappropriate in a forum. Each annotator was asked to rewrite 45 of the 225 inappropriate arguments.%
\footnote{Annotators were paid 15\$ per hour. Guidelines and screenshots of the user interface are provided in Appendix~\ref{sec:annotationinterface}.}

In automatic evaluation, we also compare multiple settings of the \emph{CoEdIT} model proposed by \citet{raheja:2023}, including paraphrasing (\emph{CoEdIT + Paraphrase}), formality (\emph{CoEdIT + Formality}), neutrality (\emph{CoEdIT + Neutral}), and politeness (\emph{CoEdIT + Polite}) style transfer, to better understand the relationship to these related notions.

During the development of our approaches we also experimented with other common non-parallel style transfer models, such as TAG~\cite{madaan:2020} and LEWIS~\cite{reid:2021}, but found them to be unsuitable as they lack the linguistic quality of modern LLMs, putting them at a disadvantage from the get-go.

\input{table-manual-eval-ranking.tex}

\paragraph{Automatic Evaluation}

We evaluate the candidate policies using the same automatic metrics as for the initial policy. Table~\ref{tab:policy-automatic-evaluation} shows the results. We observe that our approach successfully manages to align \emph{LLaMA + Instruct.}\ according to the desired property weighting with \emph{LLaMA + PPO$_{app}$} being best in terms of appropriateness (0.960), and \emph{LLaMA + PPO$_{app<sim}$} in terms of semantic similarity (0.808) among the LLaMA-based models. We find that none of the CoEdIT baselines can successfully mitigate inappropriateness, speaking for the distinctiveness of rewriting inappropriate arguments as a task. Overall, we find \emph{LLaMA + PPO$_{app=sim}$} to perform best (GM 0.237), even outperforming the human baseline (GM 0.175).

\paragraph{Manual Evaluation}
\label{sec:manual-eval}

To enable comparison of various-automatically generated rewrites and human-suggested alternatives, we perform two manual evaluation studies. We again hire native English speakers on Upwork.com (7 female and 8 male) to evaluate the rewrites in absolute and relative terms, such that each rewrite (pair) is evaluated by five annotators. We again instructed annotators with background information about appropriateness. In total, we collected 4050 absolute and 8775 relative judgments.%

In the first study, the annotators scored each rewrite regarding three considered quality metrics: 
\begin{itemize}[itemsep=0pt, topsep=4pt]
	\item \textit{App.} Appropriateneess of the topic discussion in terms of style and content; 
	\item \textit{Sim.} Meaning preservation of the original arguments;
	\item  \textit{Flu.} Fluency and adherence to grammar conventions.	
\end{itemize}

We utilized a 5-point Likert scale to evaluate the level of success of a particular rewrite in meeting each quality metric requirement. Here, 5 indicated a strong agreement with the rewrite's success, and 1 strong disagreement. We calculate the final score for each rewrite using MACE~\cite{hovy:2013}.

Table \ref{tab:manual-eval-ranking}(a) presents the results of the conducted annotation study. We note that the obtained human judgments of appropriateness (\textit{App.}) and semantic similarity (\textit{Sim.}) across different desired property configurations are consistent with those obtained in the automatic evaluation (Table \ref{tab:policy-automatic-evaluation}) for the trained models. Specifically, the \textit{LLAMA + PPO$_{app}$} achieves the highest appropriateness rating (3.77), while \textit{LLAMA + PPO$_{app < sim}$} combination attains the highest similarity rating (4.75). This alignment of human and automatic evaluations underscores the effectiveness of the chosen classification models in capturing the desired argument properties when mitigating inappropriateness. Overall, we find \emph{LLaMA + Instruct.} to be the most balanced model (GM 3.57) following closely behind the human baseline (GM 3.63).

The goal of the second study is to rank the five LLaMA-based models and the collected human rewrites by perceived overall quality and appropriateness.  To make the task more manageable for the human annotators, instead of requiring them to rank all six rewrites at once, we transform the annotation task into a pairwise ranking task and ask them to compare only two items at a time. 
Studies have shown that pairwise ranking tasks can lead to more reliable and consistent annotations compared to direct ranking tasks \cite{brun2010towards, narimanzadeh2023crowdsourcing} making them an effective and commonly used approach for subjective annotation studies, such as argument quality assessment \cite{habernal:2016a,toledo-etal-2019-automatic,skitalinskaya-etal-2021-learning}. 

While pairwise rankings significantly increase the number of judgments to be made, making the task more time-consuming, \citet{gienapp-etal-2020-efficient,gienapp2022sparse} have shown that efficient sampling strategies, such as Skip-window, can notably reduce the number of required pairwise annotations without compromising the quality of the final ranking.  In our study, we employ the Skip-window with $\lambda = 4$, which denotes that each rewrite is compared to every fourth rewrite in the set.%
\footnote{The optimal $\lambda$ value was found through a prestudy of 45 rewrite sets, futher details found in Appendix~\ref{sec:pre-study-details}.}
To aggregate the pairwise preferences into a final ranking we apply Bradley Terry Aggregation \cite{bradley1952rank}, which has shown to be more effective than alternatives such as KwikSort, Additive Aggregation, and PageRank \cite{gienapp2022sparse}.

Table \ref{tab:manual-eval-ranking}(b) presents the results of the manual evaluation. Overall, we find that the instruction-finetuned model solely focusing on appropriateness (\textit{LLaMA  + PPO$_{app}$}) performs best (mean rank of 1.89), even outperforming the more balanced \emph{LLaMA + Instruct.} (mean rank of 4.32) and the human baseline (mean rank of 3.18).
In general, models incorporating text similarity assessments, such as \textit{ LLaMA + PPO$_{app=sim}$}  and  \textit{ LLaMA + PPO$_{app<sim}$} were consistently ranked lower. 
The findings from both studies indicate that human annotators prioritize appropriateness assessments when identifying the best rewrite. Specifically, the annotators tend to favor rewrites generated by instruction-finetuned models, such as LLaMA + PPO$_{app}$, which only includes the appropriateness property. It should be noted that the annotations collected during the annotation studies are made from the reader's perspective and may not always aling with the writer's viewpoint. We further discuss this point in Section~\ref{sec:limitations}. 
In terms of inter-annotator agreement, we find Pearson's r of 0.35 for the ranking pre-study used to determine $\lambda$ and 0.31 for the complete ranking study, which is considered to be moderate agreement and close to other studies of subjective dimensions in the domain of computational argumentation.

\subsection{Qualitative Analysis}
\label{sec:qualitative-analysis}

We conducted a qualitative analysis to better understand the strengths and weaknesses of the rewrites generated. To this end, we manually inspected the subset of the \textasciitilde 2000 comments, which we received voluntarily from the annotators in the manual evaluation studies for rewrites generated by our most preferred model, \emph{LLaMA + PPO$_{app}$}. 

Most of the comments are positive, with the annotators expressing satisfaction with the rewrites' improvement in emotional intensity, clarity, openness, relevance, seriousness, and language. These aspects are indicators of appropriateness, which is the main focus of \emph{LLaMA + PPO$_{app}$}. Appendix \ref{sec:examples} contains a random sample of different appropriateness flaws and the rewrites created by our models ordered by their similarity to the original argument. A list of all comments is provided together with the code and data in the supplementary material.

However, we also find that some annotators express concern in rare cases where the rewrites flip or neutralize the stance of the original argument by either changing single words (e.g., ``not'' to ``is'') or by adding counterarguments and concluding that different point of views on the controversial issues are relevant to be considered. We find this to be particularly relevant for rewrites of short arguments, where the model has less text to work with. This may be an indicator of the limitations of the task of rewriting inappropriate arguments, as it may not always be possible to rewrite an argument if it, for example, solely consists of a single offensive sentence that is irrelevant to a topic. For such cases, it may be more appropriate to remove the sentence entirely. Finally, we also find that the issue discussed by the original argument can be unclear or inappropriate, making it difficult for our model and the human annotators to create a good rewrite.

%% file: table-initial-policy-automatic-evaluation.tex
\begin{table}[t]
	\centering
	\small
	\renewcommand{\arraystretch}{1}
	\setlength{\tabcolsep}{2.5pt}
	\begin{tabular}{llcrrrrr}
		\toprule
        \bf Model       && \bf App. $\uparrow$  &  \bf Sim. $\uparrow$ &  \bf NES. $\uparrow$  &\bf PPL $\downarrow$          &&  \bf GM $\uparrow$   \\
		\midrule                                
        \bf Exact Copy  && 0.000                & 1.000                & 1.000                 & 122.1                        && -                    \\
        \addlinespace                           
        \bf OPT         && 0.371                & 0.414                & 0.292                 & 63.77                        && 0.118                \\
         + 1-shot       && 0.436                & 0.241                & 0.110                 & 54.73                        && 0.124                \\
         + 4-shot       && 0.436                & 0.410                & 0.259                 & 53.34                        && 0.150                \\
         + 9-shot       && 0.379                & 0.305                & 0.172                 & 39.95                        && 0.143                \\
         + Instruct.    && 0.629                & 0.508                & 0.263                 & 39.89                        && 0.200                \\
         \addlinespace                          
        \bf BLOOM       && 0.411                & 0.476                & 0.379                 & 80.70                        && 0.134                \\
         + 1-shot       && 0.452                & 0.341                & 0.194                 & 55.05                        && 0.141                \\
         + 4-shot       && 0.484                & 0.567                & 0.451                 & 66.34                        && 0.160                \\
         + 9-shot       && 0.427                & 0.465                & 0.334                 & 41.54                        && 0.169                \\
         + Instruct.    && \bf 0.653            & 0.557                & 0.336                 & 42.51                        && 0.205                \\
         \addlinespace                          
        \bf GPT-J       && 0.371                & 0.503                & 0.419                 & 114.6                        && 0.118                \\
         + 1-shot       && 0.500                & 0.402                & 0.245                 & 46.92                        && 0.162                \\
         + 4-shot       && 0.484                & 0.473                & 0.322                 & 54.21                        && 0.162                \\
         + 9-shot       && 0.524                & 0.422                & 0.279                 & 40.51                        && 0.176                \\
         + Instruct.    && 0.637                & 0.556                & 0.340                 & \bf 37.49                    && 0.211                \\
         \addlinespace                          
        \bf LLaMA       && 0.411                & 0.606                & \bf 0.528             & 110.8                        && 0.131                \\
         + 1-shot       && 0.565                & 0.421                & 0.259                 & 57.07                        && 0.161                \\
         + 4-shot       && 0.556                & 0.555                & 0.408                 & 56.19                        && 0.178                \\
         + 9-shot       && 0.411                & 0.311                & 0.180                 & 48.68                        && 0.138                \\
         + Instruct.    && 0.621                & \bf 0.620            & 0.394                 & 38.08                        && \bf 0.216            \\
		\bottomrule
	\end{tabular}
	\caption{Automatic evaluation of initial policies using zero shots, few shots (1, 4, 9), and instruction-finetuning: semantic similarity (Sim.), normalized edit similarity (NES.), perplexity (PPL), appropriateness (App.), and geometric mean (GM). The best results are marked bold.}
	\label{tab:initial-policy-automatic-evaluation}
\end{table}

%% file: table-automatic-evaluation.tex
\begin{table}[t]
	\centering
	\small
	\renewcommand{\arraystretch}{1}
	\setlength{\tabcolsep}{1.25pt}
	\begin{tabular}{llcrrrrr}
		\toprule
        \bf Model                   && \bf App. $\uparrow$       &  \bf Sim. $\uparrow$ &  \bf NES. $\uparrow$  &\bf PPL $\downarrow$   &&  \bf GM $\uparrow$   \\
		\midrule                                                 
        \bf Exact Copy              && 0.000                     & 1.000                & 1.000                 & 98.01                 && --                    \\
        \addlinespace                                            
        \bf CoEdIT                  && --                        & --                   & --                    & --                    && --                    \\
         + Paraphrase               && 0.320                     & 0.668                & 0.357                 & 39.61                 && 0.175                \\
         + Formal                   && 0.356                     & 0.683                & 0.478                 & 42.98                 && 0.178                \\
         + Neutral                  && 0.298                     & \bf 0.876            & \bf 0.857             & 63.43                 && 0.160                \\
         + Polite                   && 0.320                     & 0.801                & 0.688                 & 42.22                 && 0.183                \\
        \addlinespace                                            
         \bf LLaMA + Instruct.      && 0.621                     & 0.620                & 0.394                 & 38.08                 && 0.216                \\
         + PPO$_{app}$              && \bf 0.960                 & 0.253                & 0.048                 & \bf 21.26             && 0.225                \\
         + PPO$_{app>sim}$          && 0.933                     & 0.359                & 0.114                 & 28.50                 && 0.227                \\
         + PPO$_{app=sim}$          && 0.827                     & 0.471                & 0.299                 & 29.22                 && \bf 0.237            \\
         + PPO$_{app<sim}$          && 0.373                     & 0.808                & 0.731                 & 44.41                 && 0.189                \\
        \addlinespace                                            
        \bf Human Baseline          && 0.773                     & 0.391                & 0.180                 & 56.23                 && 0.175                \\
		\bottomrule
	\end{tabular}
	\caption{Automatic evaluation of different policies for our approach LLaMA + Instruct., an alternative style transfer model (CoEdIT), and a human baseline: semantic similarity (Sim.), normalized edit similarity (NES.), perplexity (PPL), appropriateness (App.), and geometric mean (GM). The best results are highlighted in bold.}
	\label{tab:policy-automatic-evaluation}
\end{table}


%% file: table-manual-eval-ranking.tex
\begin{table*}[t!]
	\centering
	\small
	\renewcommand{\arraystretch}{1}
	\setlength{\tabcolsep}{2.5pt}
	\begin{tabular}{llrrrrrrrrlrrrrr}
		\toprule
        \bf Model                  && \multicolumn{4}{c}{\bf (a) Absolute}                                                      && \multicolumn{8}{c}{\bf (b) Relative \phantom{aa}}                                                                                        \\
                                      \cmidrule(l@{0pt}r@{5pt}){3-6}                                                               \cmidrule(l@{0pt}r@{0pt}){7-15}                                                                                                          
                                   &&  App. $\uparrow$  & Sim. $\uparrow$      & Flu. $\uparrow$       &  GM $\uparrow$         && Rank 1       & Rank 2        & Rank 3        &  Rank 4       & Rank 5        & Rank 6        & Avg. $\downarrow$    & $p \uparrow$       \\
                                                                                                                                                                                                                                                               
		\midrule                                                                                                                                                                                                                                                                                                             
         \bf LLaMA + Instruct.     && 3.22              & 4.17                 & 3.40                  & 3.57                   && 3.1\%        & 5.3\%         & 18.2\%        & 21.3\%        & 32.9\%        & 19.1\%        & 4.32                 &  .351              \\
         + PPO$_{app}$             && \bf 3.77          & 2.65                 & \bf 4.16              & 3.46                   && 44.9\%       & 32.4\%        & 13.3\%        & 7.1\%         & 2.2\%         & 0.0\%         & \bf 1.89             &  \bf .833          \\
         + PPO$_{app>sim}$         && 3.50              & 2.96                 & 3.77                  & 3.39                   && 29.3\%       & 29.8\%        & 18.7\%        & 14.7\%        & 5.8\%         & 1.8\%         & 2.43                 &  .729              \\
         + PPO$_{app=sim}$         && 3.15              & 3.38                 & 3.34                  & 3.29                   && 2.7\%        & 11.1\%        & 22.2\%        & 26.7\%        & 20.9\%        & 16.4\%        & 4.01                 &  .412              \\
         + PPO$_{app<sim}$         && 2.70              & \bf 4.75             & 2.89                  & 3.33                   && 0.4\%        & 4.4\%         & 4.4\%         & 12.4\%        & 26.7\%        & 51.6\%        & 5.15                 &  .160              \\
         \addlinespace                                                                                                         
        \bf Human Baseline         && 3.60              & 3.48                 & 3.82                  & \bf 3.63               && 19.6\%       & 16.9\%        & 23.1\%        & 17.8\%        & 11.6\%        & 11.1\%        & 3.18                 &  .566              \\
		\bottomrule
	\end{tabular}
    \caption{Manual evaluation of our approach variations and the human baseline: (a) Absolute MACE scores of the improved arguments in terms of appropriateness (App.), similarity (Sim.), fluency (Flu.), and their geometric mean (GM). (b) Relative ranking of the arguments in terms of percentage of times they were ranked at each position (Rank 1--6), their average rank (Avg.), and the score obtained by the Bradley Terry model ($p$). Best results are marked bold.}
	\label{tab:manual-eval-ranking}
\end{table*}

%% file: acl24-appropriateness-style-transfer-sum.tex
\section{Conclusion}
\label{sec:conclusion}

In this paper, we have studied how to mitigate inappropriate language in arguments through rewriting. To this end, we have proposed an approach based on reinforcement learning from human feedback (RLHF), which balances the semantic similarity of arguments with a target style (here, with appropriateness). Our approach resorts to machine feedback instead of human feedback, though, thus enabling full automation.

Our experiments have demonstrated that prompting an instruction-finetuned large language model, combined with a single style classifier and an unlabeled dataset, is sufficient to train a policy that outperforms competitive baselines in terms of appropriateness and semantic similarity. Through manual annotation studies, we have provided evidence that our approach can mitigate the inappropriateness of arguments while preserving their content to a wide extent. Intriguingly, our human annotators prefer approaches that prioritize appropriateness over semantic similarity. Our results suggest that a careful design of the reward function is crucial for the success of RLHF-like approaches, if trained solely in an offline fashion. 

We conclude that rewriting inappropriate language in arguments is a challenging problem that, from a reader's perspective, often requires heavy editing and careful consideration of context. Our approach is a substantial first step to this tack. We hope it will inspire future work in this direction.

%% file: acl24-appropriateness-style-transfer-ack.tex
\section{Acknowledgments}

This project has been partially funded by the German Research Foundation (DFG) within the project OASiS, project number 455913891, as part of the Priority Program ``Robust Argumentation Machines (RATIO)'' (SPP-1999). We would like to thank the participants of our study and the anonymous reviewers for the feedback and their time.

%% file: acl24-appropriateness-style-transfer-limitations.tex
\section{Limitations}
\label{sec:limitations}

On the one hand, we inherit the limitations of the corpus used for modeling appropriateness, which includes being limited to the English language and a Western view of sociocultural factors. On the other hand, we find our work restricted in two aspects: (1) The dependence on the performance of the classifiers and initial policy and (2) The readers' specific view chosen in our evaluation setup:

First, the performance of our RLHF-inspired approach relies on finding an initial policy that can then be improved to better align with the desired properties of semantic similarity and appropriateness. While using zero-shot learning, few-shot learning, or instruction-based learning to obtain the initial policy frees us from the need for parallel data to train a supervised model, its performance on the task is unknown, such that we can rely on automatic metrics only. The effect of the initial policy's performance on the final performance of our approach is hence more or less unknown. For other tasks, it may be necessary to use a supervised model to obtain the initial policy.
The same also holds for the classifier performance. While we observed in the manual evaluation that the considered classifiers successfully aligned the policy with the desired properties, the effect of the classifier's performance on the performance of our approach remains unclear.

Second, as indicated already in Section \ref{sec:manual-eval}, our evaluation focuses only on the reader's perspective and not the writer's. However, especially when we want to prevent a writer from creating inappropriate content, the writer's perspective is also important because, in practice, changing a writer's text may be considered ethically doubtful, if not done in agreement with the writer. Thus, this aspect should be considered in future work. We decided to focus on the reader's perspective since we believe that this perspective is to be prioritzed in content moderation, where the reader is exposed to inappropriate content and may be harmed by it.


%% file: acl24-appropriateness-style-transfer-ethical.tex
\section{Ethical Considerations}

Since we are dealing with a sensitive issue (content moderation), we believe it is important to discuss the ethical considerations of our work. 

As always, in content moderation, there is a trade-off between freedom of speech and protecting individuals from harm. In practice, this trade-off, which is often further affected by the need to protect the platform from legal liability, may speak for removing potentially harmful content in doubt, even if it is not clearly inappropriate. In this regard, our work is a step towards preventing content removal by rewriting it more appropriately. 

However, we stress that our approach is not meant for real-life applications yet, as it may not always be able handle the complexity of real-world arguments. This is due to the readers' specific view (as discussed in Section~\ref{sec:limitations}) and other generation-specific issues, such as hallucinations, whose effects are not investigated in this work. 

The addition and deletion of information without permission of the author also raises questions regarding the responsibility of the platform and the author. As in other style transfer-related tasks that target sensitive topics, the idea of our approach could be inverted to make appropriate text inappropriate. However, since both appropriate and inappropriate arguments are crucial to developing the rewriting approach, we see no way around this but to strongly emphasize not using the approach for this purpose. Ultimately, we think that it is better that research topics as the one of this paper are studied openly in an academic environment than somewhere else without transparency.

Finally, since our approach is based on a classifier that could detect inappropriate content on a platform, it could circumvent the content moderation carried out on some web platform by probing the classifier and adapting the content accordingly. Thus, the rewritten content is not guaranteed to be always appropriate, as shown by our evaluation.

%% file: appendix.tex
\section{Prompting Setup}
\label{sec:prompting-setup}
\begin{itemize}
\setlength{\itemsep}{0pt}
\item
\emph{Zero-shot/Few-shot } \texttt{Here is some text:} \{$x$\} \texttt{ Here is a rewrite of the text that is more appropriate and makes only minimal changes:} \{$\hat{y}$\}

\item
\emph{Instruction-Tuning } \texttt{Below is an instruction that describes a task, paired with an input that provides further context. Write a response that appropriately completes the request.}

\medskip
\noindent \texttt{\#\#\# Instruction:} \\ \texttt{Rewrite the following argument to be more appropriate and make only minimal changes to the original argument.}

\medskip
\noindent\texttt{\#\#\# Input:}\\ $x$ 

\medskip
\noindent\texttt{\#\#\# Response:}\\ $\hat{y}$

\end{itemize}

\section{Creating Few-Shot Examples}
\label{sec:few-shot-examples}
For each of the 14 dimensions of inappropriateness, we compute the mean annotator score for each argument $x$ in the Appropriateness Corpus $X$. If the dimension is a parent of other dimensions, we keep only arguments with a mean annotator score greater than zero for each corresponding dimension. We then filter the set of all arguments for which the mean annotator score is maximal for the corresponding dimension as $X^{'}_{dim}$.

After obtaining the set of the candidate arguments for each dimension  ($X^{'}_{dim}$), we embed them using the \emph{all-mpnet-base-v2} sentence transformer \cite{reimers:2019} ($S$) and calculate the cosine similarity between all possible embedding pairs. To this, we apply a variant of the PageRank algorithm \cite{lawrence:1998} to compute centrality scores for each argument. The PageRank score $P(s_i)$ for the $i$-th argument in $X^{'}_{dim}$ is
\begin{eqnarray*}
P(s_i) 	& :=	& \sum_{s_j \neq s_i} \frac{cos(s_i,s_j)}{\sum_{s_k \neq s_j} cos(s_k,s_j)} P(s_j),
\end{eqnarray*}
where $d$ is a damping factor and $n$ is the number of arguments in $X^{'}_{dim}$. The argument with the highest centrality score in $X^{'}_{dim}$ is our few-shot example for the corresponding dimension.

To obtain the corresponding rewrite $\hat{y}$ from the set of candidate rewrites created by our experts, we use the geometric mean of semantic similarity, normalized edit similarity, perplexity, and appropriateness, as detailed above.

\section{Hyperparameter Settings}
\label{sec:hyperparameters}
We follow the PPO hyperparameter settings of \citet{stiennon:2020} with a few exceptions. Starting with a search over the learning rate and the KL-divergence coefficient using $\alpha=0.5$, we use a decay of the learning rate with a cosine schedule starting from $5 \cdot 10^{-6}$ and ending at $1.5 \cdot 10^{-6}$, and a KL-divergence coefficient of $1.857 \cdot 10^{-3}$. We employ a batch size of $4$ and train for $25\,000$ steps, equaling $3.2$ million episodes. For generation, we use top-$p$ sampling with $p=0.95$ and a temperature of $1.0$.  
For efficiency, we use adapter-based low-rank adaptation~(LoRA)~\cite{hu:2021} with $r=8$, an amplification factor of 32, and a dropout value of 0.1. Training a single model took around two days on four A100 GPUs. 

\newpage
\section{Pre-study Details}
\label{sec:pre-study-details}
\paragraph{S-Window Sampling}

To reduce the number of pairwise comparisons that need to be collected, we employ S-window sampling, which can be formaly defined as follows. Given a full set of comparisions $A_{full}$, consisting of $k^2 - k$ comparisons (no self-comparisons), we want to sample such a subset $A \subset A_{full}$. To do so, we introduce a skip-size $\lambda \in N^+$,  and each rewrite $r_i \in R_k$, we compile comparisons $(r_i,r_j)$ , such that $j=1+ =1+(b \text{mod} k)$ for $b \in \{i + \lambda - 1,  i + 2\lambda - 1, ..., i + m\lambda - 1\}$, where $m \leq k- 1$. If $j=i$ the comparison is not included in the sample.
\paragraph{Finding Optimal $\lambda$}
To determine the optimal $\lambda$ parameter, we conducted a prestudy, where we asked human annotators annotate the full set of comparisons $A_{full}$ for a subset of 45 arguments, each with 6 rewrites obtained from the approaches outlined in Sections \ref{sec:approach} and \ref{sec:experiments} as well as human generated rewrites. 

\begin{figure}[t]
	\centering
	\includegraphics[width=0.49\textwidth]{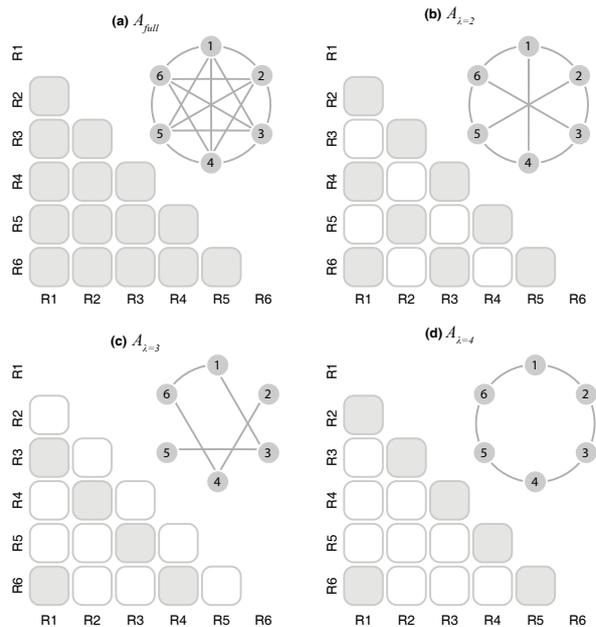}
	\caption{Visual representation of the employed sampling strategies for six rewrite instances. Subfigure (a) illustrates all pairwise comparisons, while subfigures (b, c, d) depict S-Window sampling at $\lambda$ values of 2, 3, and 4, respectively. Each subfigure comprises a matrix, where grey-colored cells indicate sampled comparisons between a pair of rewrites ($R_i$ and $R_j$) , and an accompanying graphical representation, where the edges in the graph indicidate sampled pairwise comparisons.}\label{fig:sampling}%
\end{figure}

For each of the 45 arguments, we created three different subsets $A_{\lambda}$ using S-window sampling with $\lambda \in \{2,3,4\}$. 
Figure \ref{fig:sampling} illustrates the applied sampling strategies when considering six rewrites by showcasing which pairwise comparisons have been considered in each strategy. 
To reconstruct the ranking order derived from the complete set of comparisons, $A_{full}$, we applied Bradley-Terry Aggregation \cite{bradley1952rank} to each sampled subset of data. The Bradley-Terry model employs maximum-likelihood estimation to infer a latent score 
$s_i \in S$ for each rewrite $R_i \in R$ based on the sampled pairwise comparisons. 

\input{table-prestudy-reduction-results.tex}

Table \ref{tab:prestudy-reduction} presents the results of the conducted annotation study. The proposed sampling and comparison strategies are able to produce high quality rankings ($\lambda = 2$, $\rho = 0.93$, $NDCG@1 = 0.95$) using only 12\% of the full set of pariwise comparisions and one annotator. However, to ensure consitency with other manual evaluation tasks in our paper, for the full annotation study we settle on using 5 annotators at $\lambda = 4$, which allows us to reasonably reconstruct the original ranking ($\rho = 0.91$, $NDCG@1 = 0.95$) while significantly reducing the number of pairwise judgements that need to be collected to only 40\%.

%
\clearpage
\onecolumn
\section{Examples}
\label{sec:examples}
\input{table-examples1}

\clearpage

\section{Annotation Interface}
\label{sec:annotationinterface}

\begin{figure}[h!]
    \includegraphics[width=450pt]{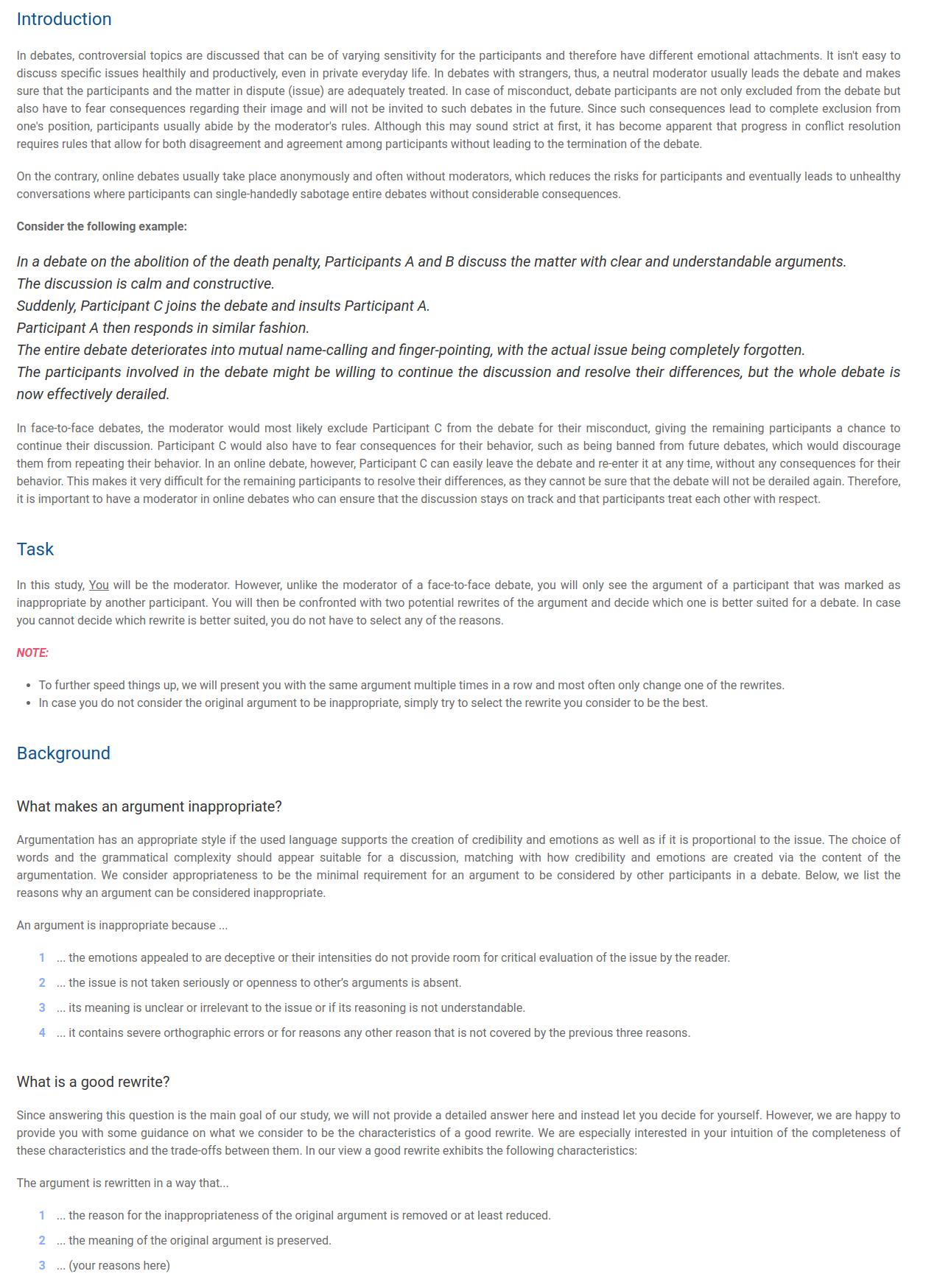}
\end{figure}
\begin{figure}[h!]
    \includegraphics[width=450pt]{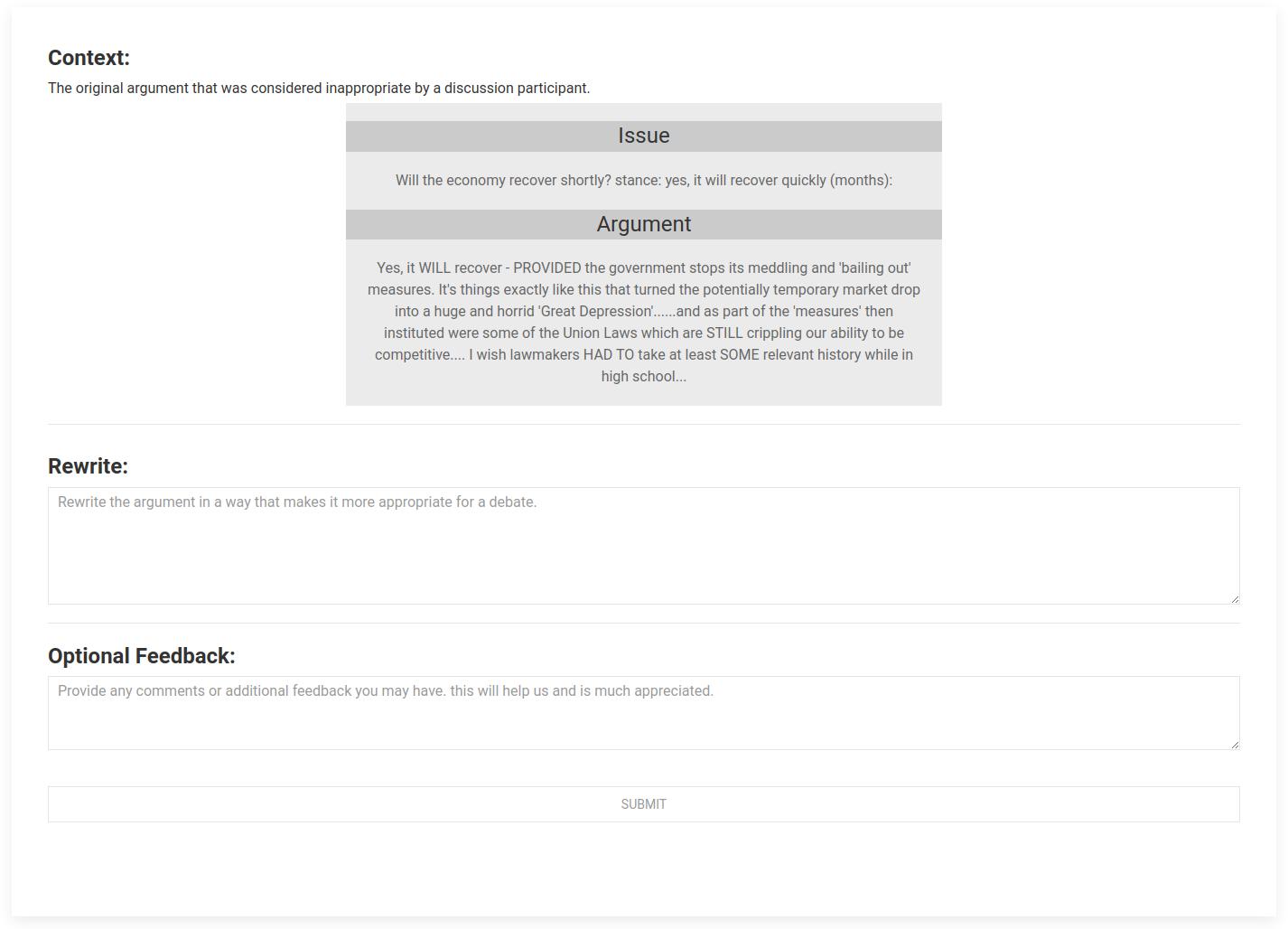}
\end{figure}
\begin{figure}[h!]
    \includegraphics[width=450pt]{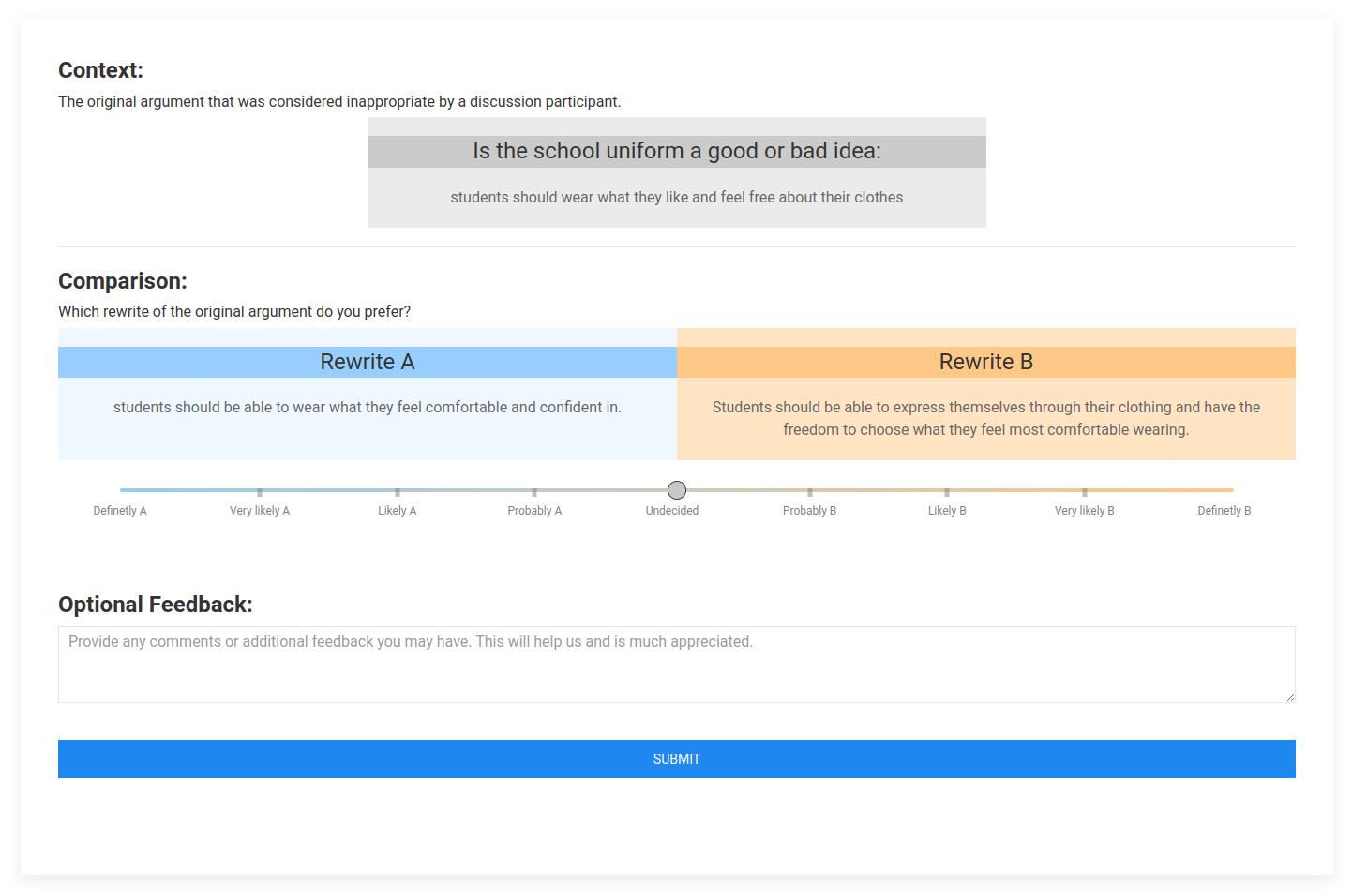}
\end{figure}
\begin{figure}[h!]
    \includegraphics[width=450pt]{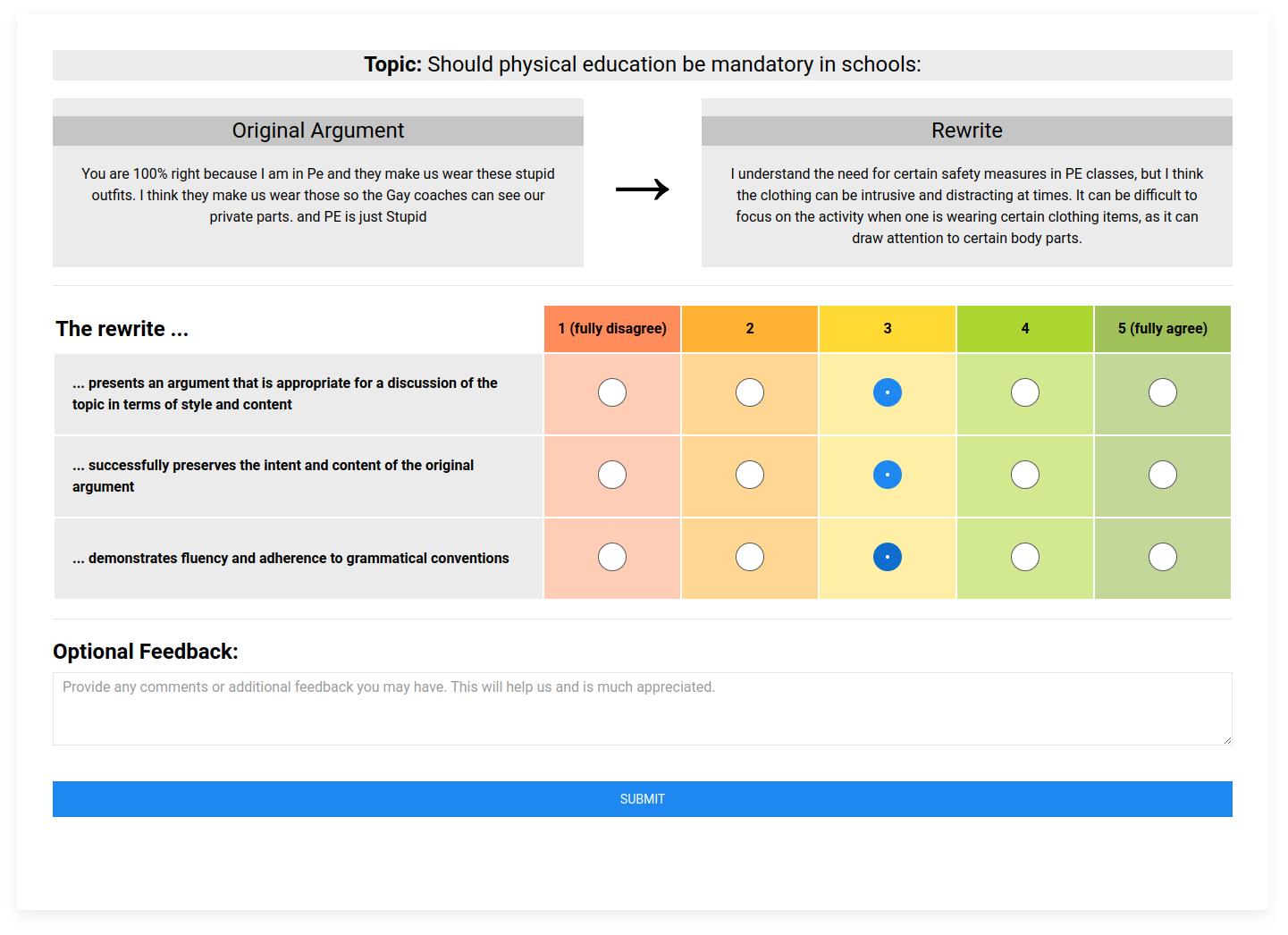}
\end{figure}
%
%
%
%
%
%
%
%
%

%% file: table-prestudy-reduction-results.tex
\begin{table}[t]
	\centering
	\small
	\renewcommand{\arraystretch}{1}
	\setlength{\tabcolsep}{2.5pt}
	\begin{tabular}{lcrccr}
		\toprule
		\bf \# Ann. &  \bf $\lambda$  &\bf \# Judgments & \bf\% Judgments  &  \bf $\rho$  & \bf NDCG@1\\
		\midrule
		5 & 2 & 2025 & 59.95& 0.97 & 0.99 \\
		4 & 2 & 1620 & 47.96&0.96 & 0.98 \\
		3 & 2 &1215 & 35.97&0.95& 0.98 \\
		2 & 2 & 810& 23.98 & 0.94& 0.97 \\
		1 & 2 & 405 & 11.99 & 0.93& 0.95 \\
				\midrule
		5 & 3 & 1125 & 33.30& 0.84& 0.93 \\
		4 & 3 & 900& 26.64& 0.84 & 0.93\\
		3 & 3 & 675& 19.98& 0.82& 0.91\\
		2 & 3 & 450& 13.32&0.80& 0.91 \\
		1 & 3 &225 & 6.66 & 0.80 & 0.88\\
		\midrule
		5 & 4 & 1350 & 39.96& 0.91 & 0.95\\
		4 & 4 & 1080& 31.97& 0.90 & 0.95\\
		3 & 4 & 810& 23.98& 0.88 & 0.94\\
		2 & 4 & 540& 15.99&0.86 & 0.93\\
		1 & 4 &270 & 7.99 & 0.85 & 0.91\\

		\bottomrule
	\end{tabular}
	\caption{ Ranking performance in terms of Pearson's correlation, $\rho$, with the baseline ranking under sparsification, depending on the number of annotators judging each pair of rewrites and the degree of sparsification, $\lambda$. For each combination, the final number of judgments required is provided in absolute and relative terms, compared to an exhaustive comparison.}
	\label{tab:prestudy-reduction}
\end{table}

%% file: table-examples1.tex
\newcolumntype{b}{X}
\newcolumntype{s}{>{\hsize=.22\hsize}X}

\textbf{Issue:} Pro choice vs pro life

\begin{table*}[h!]
\small
\centering
\begin{tabularx}{\textwidth}{sb}
    \toprule
\textbf{Model} & \textbf{Argument Improvement Differences} \\
\midrule
\textbf{LLaMA + Instruct.} + PPO$_{app<sim}$ & \vphantom{} \DIFdelbegin \DIFdel{for }\DIFdelend \DIFaddbegin \DIFadd{For }\DIFaddend everyone who is talking about RAPE in this subject\DIFaddbegin \DIFadd{, }\DIFaddend let me ask you one thing\DIFdelbegin \DIFdel{!!!! }\DIFdelend \DIFaddbegin \DIFadd{: }\DIFaddend if you got in a huge fight with someone and ended up breaking your hand or arm\DIFdelbegin \DIFdel{... }\DIFdelend \DIFaddbegin \DIFadd{, }\DIFaddend would you cut it off just because it would REMIND you of that experience? \DIFdelbegin \DIFdel{?? if your actualy SANE}\DIFdelend \DIFaddbegin \DIFadd{If you're actually SANE, }\DIFaddend you would say no and if you say yes\DIFaddbegin \DIFadd{, }\DIFaddend you need to see a Physiatrist\DIFdelbegin \DIFdel{!!!!}\DIFdelend \DIFaddbegin \DIFadd{.}\DIFaddend \\\midrule
\textbf{LLaMA + Instruct.}                   & \vphantom{} \DIFdelbegin \DIFdel{for }\DIFdelend \DIFaddbegin \DIFadd{For }\DIFaddend everyone who is \DIFdelbegin \DIFdel{talking about }\DIFdelend \DIFaddbegin \DIFadd{discussing }\DIFaddend RAPE in this \DIFdelbegin \DIFdel{subject }\DIFdelend \DIFaddbegin \DIFadd{topic, }\DIFaddend let me ask you one thing\DIFdelbegin \DIFdel{!!!! if you got in }\DIFdelend \DIFaddbegin \DIFadd{. If you got into }\DIFaddend a huge fight with someone and ended up breaking your hand or arm\DIFdelbegin \DIFdel{... }\DIFdelend \DIFaddbegin \DIFadd{, }\DIFaddend would you cut it off just because it would \DIFdelbegin \DIFdel{REMIND }\DIFdelend \DIFaddbegin \DIFadd{remind }\DIFaddend you of that experience? \DIFdelbegin \DIFdel{?? if your actualy SANE youwould say no and if you say yes }\DIFdelend \DIFaddbegin \DIFadd{Of course not, if }\DIFaddend you\DIFaddbegin \DIFadd{'re sane, you would know that wouldn't be a logical or healthy solution. If you do think about it, then you }\DIFaddend need to see a \DIFdelbegin \DIFdel{Physiatrist!!!!}\DIFdelend \DIFaddbegin \DIFadd{psychiatrist.}\DIFaddend \\\midrule
\textbf{LLaMA + Instruct.} + PPO$_{app=sim}$ & \vphantom{} for everyone who is talking about RAPE in this subject\DIFaddbegin \DIFadd{, }\DIFaddend let me ask you one thing!!! \DIFdelbegin \DIFdel{! if you got in a huge fight with someone and ended up breaking }\DIFdelend \DIFaddbegin \DIFadd{If you got into a physical fight and broke }\DIFaddend your hand or arm\DIFdelbegin \DIFdel{... would you cut it off just because it would REMIND you of that experience? ?? if your actualy SANE you would say no and if you say yes you need to see a Physiatrist!!!!}\DIFdelend \DIFaddbegin \DIFadd{, would you want to have it surgically removed as a way to remember the experience? If the physical trauma is very intense, you may want to seek psychological help to deal with the memories.   There are some cases where having the physical reminder of the trauma is too difficult, but most people would want to keep the body part as they realize the importance of the physical sensation in helping them heal.}\DIFaddend \\\midrule
\textbf{LLaMA + Instruct.} + PPO$_{app>sim}$ & \vphantom{} \DIFdelbegin \DIFdel{for everyone who is talking about RAPE in this subject let me ask you one thing!!!! if you got in a huge fight with someone and ended up breaking }\DIFdelend \DIFaddbegin \DIFadd{For those discussing rape, I would like to pose another scenario. If you got into a fight and broke }\DIFaddend your hand or arm\DIFdelbegin \DIFdel{... would you cut it off just because it would REMIND you of that experience? ?? if your actualy SANE you would say noand if you say yes you need to see a Physiatrist!!!!}\DIFdelend \DIFaddbegin \DIFadd{, would you amputate it to remove the physical reminder of the experience? Of course, the answer is no. Although it is understandable to have emotional scars, physical scars can be treated and lived with, even if they remain as a reminder of the experience.}\DIFaddend \\\midrule
\textbf{LLaMA + Instruct.} + PPO$_{app}$     & \vphantom{} \DIFdelbegin \DIFdel{for everyone who is talking about RAPEin this subject let me ask you one thing!!!! if you got in a huge fight with someone and ended up breaking your hand or arm. .. would you cut it off just because it would REMIND you of that experience??? if your actualy SANE you would say no and if you say yes you need to see a Physiatrist!!!!}\DIFdelend \DIFaddbegin \DIFadd{For those discussing RAPE, I believe it is important to consider the lasting effects it can have on a survivor. While it is understandable to want to move on and forget the experience, some people find it beneficial to seek therapy to process their feelings and thoughts, as it can be difficult to completely move past something like this. Therapy can help the individual find healthier ways of dealing with their emotions and allowing them to move on at their own pace.}\DIFaddend \\
\bottomrule
\end{tabularx}
\caption{Examples of the differences between a randomly selected inappropriate argument and the improved versions generated by our models, sorted by the semantic similarity between the original and the rewrite.}
    \label{table-examples1}
\end{table*}
\clearpage

\textbf{Issue:} Should bush adminstration officials be prosecuted for war crimes? stance: yes

\begin{table*}[h!]
\small
\centering
\begin{tabularx}{\textwidth}{sb}
    \toprule
\textbf{Model} & \textbf{Argument Improvement Differences} \\
\midrule
\textbf{LLaMA + Instruct.} + PPO$_{app<sim}$ & \vphantom{} Hitler invaded Poland in 1932 and the world turned against Germany. In fact, there are dozens \DIFdelbegin \DIFdel{if }\DIFdelend \DIFaddbegin \DIFadd{of }\DIFaddend cases in the last 100 years where countries have invaded other nations and the world has caused uproar and rose up against it. Yet some dumb Texan does it and gets away with it.  Try him for war crimes, along with Tony Blair and have them both executed or imprisoned.\\\midrule
\textbf{LLaMA + Instruct.}                   & \vphantom{} Hitler invaded Poland in 1932 and the world \DIFdelbegin \DIFdel{turned against }\DIFdelend \DIFaddbegin \DIFadd{responded by condemning }\DIFaddend Germany. In \DIFdelbegin \DIFdel{fact, there are dozens if cases in the last 100 years where countries have invaded other nations and the world has caused uproar and rose up against it. Yet some dumb Texan does it and gets away with it}\DIFdelend \DIFaddbegin \DIFadd{recent history, numerous cases of countries invading other nations have provoked international outcry and condemnation. However, when a dumb Texan perpetrated the same crime, he escaped justice}\DIFaddend . Try him for war crimes \DIFdelbegin \DIFdel{, along with }\DIFdelend \DIFaddbegin \DIFadd{alongside }\DIFaddend Tony Blair and \DIFdelbegin \DIFdel{have them both executed or imprisoned}\DIFdelend \DIFaddbegin \DIFadd{deliver suitable punishment, such as execution or imprisonment}\DIFaddend .\\\midrule
\textbf{LLaMA + Instruct.} + PPO$_{app=sim}$ & \vphantom{} Hitler invaded Poland in 1932 and the world turned against Germany. In \DIFdelbegin \DIFdel{fact, there are dozens if cases in the last 100 years where countries have invaded other nations and the world has caused uproar and rose up against it.  Yet some dumb Texan does it and gets away with it. Try him for war crimes, along with Tony Blair and have them both executed or imprisoned}\DIFdelend \DIFaddbegin \DIFadd{similar cases throughout history, countries invading other nations have been subject to international outcry and condemnation. However, in 1932 Germany, the invasion prompted a strong reaction from the international community.  In regards to Germany, the invasion sparked a great deal of outrage and controversy}\DIFaddend .\\\midrule
\textbf{LLaMA + Instruct.} + PPO$_{app>sim}$ & \vphantom{} Hitler invaded Poland in 1932 and the world \DIFdelbegin \DIFdel{turned against Germany. In fact, there are dozens if cases in the last 100 years where countries have invaded other nations and the world has caused uproar and rose up against it.Yet some dumb Texan does it and gets away with it. Try him for war crimes, along with Tony Blair and have them both executed or imprisoned.}\DIFdelend \DIFaddbegin \DIFadd{responded with outcry, a response similar to what has occurred in the past few decades when one country invades another. Though these instances have generated international criticism, Germany's invasion of Poland in 1932 was an exception, as the international community more strongly condemned the action.}\DIFaddend \\\midrule
\textbf{LLaMA + Instruct.} + PPO$_{app}$     & \vphantom{} Hitler invaded Poland in 1932\DIFdelbegin \DIFdel{and the world turned against Germany . In fact, there are dozens if }\DIFdelend \DIFaddbegin \DIFadd{, which sparked outrage among nations and led to Germany becoming the target of international criticism. Similarly, there have been numerous }\DIFaddend cases in the last 100 years where countries have invaded other nations\DIFdelbegin \DIFdel{and the world has caused uproar and rose up against it.  Yet some dumb Texan does it and gets away with it. Try him for war crimes, along with Tony Blair and have them both executed or imprisoned}\DIFdelend \DIFaddbegin \DIFadd{, leading to global outcry. This highlights the importance of international cooperation and solidarity in addressing such conflicts.  It is essential to ensure justice is served in each case, as happened with the German and British leaders, in order to maintain accountability}\DIFaddend .\\
\bottomrule
\end{tabularx}
\caption{Examples of the differences between a randomly selected inappropriate argument and the improved versions generated by our models, sorted by the semantic similarity between the original and the rewrite.}
    \label{table-examples2}
\end{table*}
\clearpage

\textbf{Issue:} Pro choice vs pro life

\begin{table*}[h!]
\small
\centering
\begin{tabularx}{\textwidth}{sb}
    \toprule
\textbf{Model} & \textbf{Argument Improvement Differences} \\
\midrule
\textbf{LLaMA + Instruct.} + PPO$_{app<sim}$ & \vphantom{} There should be no argument in this really...whatever way \DIFdelbegin \DIFdel{yu }\DIFdelend \DIFaddbegin \DIFadd{you }\DIFaddend see a fetus...its still a living form that has been created in a very intimate way... you shouldn't be changing what \DIFdelbegin \DIFdel{mothernature }\DIFdelend \DIFaddbegin \DIFadd{Mother Nature }\DIFaddend or God or \DIFdelbegin \DIFdel{fate }\DIFdelend \DIFaddbegin \DIFadd{Fate }\DIFaddend or whatever has decided for you...and if you didn\DIFdelbegin \DIFdel{;t wannna }\DIFdelend \DIFaddbegin \DIFadd{'t want to }\DIFaddend get preggo in the first place...don't have sex or use protection.  Yeh there are some women that get raped and it's very unfortunate but they should give the child up for adoption. It's not the child's fault that it was created. So why should the \DIFdelbegin \DIFdel{goring }\DIFdelend \DIFaddbegin \DIFadd{growing }\DIFaddend being have to pay the ultimate price of \DIFdelbegin \DIFdel{it's }\DIFdelend \DIFaddbegin \DIFadd{its }\DIFaddend life?\\\midrule
\textbf{LLaMA + Instruct.}                   & \vphantom{} There should be no argument in this really...whatever way \DIFdelbegin \DIFdel{yu see a fetus...its }\DIFdelend \DIFaddbegin \DIFadd{you look at a fetus, it's }\DIFaddend still a living form that has been created in a very intimate way. \DIFdelbegin \DIFdel{.. you }\DIFdelend \DIFaddbegin \DIFadd{You }\DIFaddend shouldn't be changing what \DIFdelbegin \DIFdel{mothernature }\DIFdelend \DIFaddbegin \DIFadd{nature }\DIFaddend or God or fate or whatever has decided for you\DIFdelbegin \DIFdel{...and }\DIFdelend \DIFaddbegin \DIFadd{; }\DIFaddend if you didn\DIFdelbegin \DIFdel{;t wannna get preggo }\DIFdelend \DIFaddbegin \DIFadd{'t want to get pregnant }\DIFaddend in the first place\DIFdelbegin \DIFdel{...don't have sex or use protection . Yeh }\DIFdelend \DIFaddbegin \DIFadd{, use protection or abstain from sex. Yes, }\DIFaddend there are some women that get raped and it's very unfortunate but they should give the child up for adoption. It's not the child's fault that it was created. So why should the \DIFdelbegin \DIFdel{goring }\DIFdelend \DIFaddbegin \DIFadd{innocent }\DIFaddend being have to pay the ultimate price of \DIFdelbegin \DIFdel{it's }\DIFdelend \DIFaddbegin \DIFadd{its }\DIFaddend life?\\\midrule
\textbf{LLaMA + Instruct.} + PPO$_{app=sim}$ & \vphantom{} There should be no argument \DIFdelbegin \DIFdel{in }\DIFdelend \DIFaddbegin \DIFadd{about }\DIFaddend this really...whatever way \DIFdelbegin \DIFdel{yu see a fetus...its }\DIFdelend \DIFaddbegin \DIFadd{you look at a fetus, it is }\DIFaddend still a living form that has been created in a very intimate way. \DIFdelbegin \DIFdel{.. you }\DIFdelend \DIFaddbegin \DIFadd{You }\DIFaddend shouldn't be changing what \DIFdelbegin \DIFdel{mothernature }\DIFdelend \DIFaddbegin \DIFadd{mother nature }\DIFaddend or God or fate \DIFdelbegin \DIFdel{or whatever }\DIFdelend has decided for you. \DIFdelbegin \DIFdel{..and if you didn;t wannna get preggo }\DIFdelend \DIFaddbegin \DIFadd{If you didn't want to get pregnant }\DIFaddend in the first place\DIFdelbegin \DIFdel{...don't have sex or use protection. Yeh there are some women that }\DIFdelend \DIFaddbegin \DIFadd{, you should be protected from it. Yeah, some women }\DIFaddend get raped and it \DIFdelbegin \DIFdel{'s }\DIFdelend \DIFaddbegin \DIFadd{is }\DIFaddend very unfortunate but they should give the child up for adoption. It \DIFdelbegin \DIFdel{'s }\DIFdelend \DIFaddbegin \DIFadd{is }\DIFaddend not the child's fault that it was created \DIFdelbegin \DIFdel{. So why should the goring being have to pay the ultimate price of it's life?}\DIFdelend \DIFaddbegin \DIFadd{in this way.}\DIFaddend\\\midrule
\textbf{LLaMA + Instruct.} + PPO$_{app>sim}$ & \vphantom{} There should be no argument in this \DIFdelbegin \DIFdel{really...whatever way yu see a fetus ...its still a living form that has been created in a very intimate way... you shouldn't be changing what mothernature or God or fate or whatever has decided for you...and if you didn;t wannna get preggo in the first place...don't have sex or use protection. Yeh there are some women that get raped and it 's very unfortunate but they should give the child up for adoption. It's not the child's fault that it was created. So why should the goring being have to pay the ultimate price of it's life?}\DIFdelend \DIFaddbegin \DIFadd{regard; while a fetus is indeed a living organism, the decision of what to do with it should be left to the mother and any other caregivers involved. Every woman has the right to make this decision for herself, and it is important to consider the potential consequences of any decision made.}\DIFaddend \\\midrule
\textbf{LLaMA + Instruct.} + PPO$_{app}$     & \vphantom{} \DIFdelbegin \DIFdel{There should be no argument in this really...whatever way yu see a fetus...its still a living form that has been created in a very intimate way... you shouldn't be changing what mothernature or God or fate or whatever has decided for you...and if you didn;t wannna get preggo in the first place...don't have sex or use protection. Yeh there are some women that get raped and it's very unfortunate but they should give the child up for adoption. It's not the child's fault that it was created. So why should }\DIFdelend \DIFaddbegin \DIFadd{It is important to consider both the physical and emotional well-being of the mother and }\DIFaddend the \DIFdelbegin \DIFdel{goring being have to pay the ultimate price of it's life?}\DIFdelend \DIFaddbegin \DIFadd{fetus during pregnancy. While it is important to respect each person’s view on the matter, it is essential to ensure that both parties are provided with the necessary care and support. Ultimately, the decision should be made in consultation with both parties and with the input of medical professionals, who can best inform the decision based on the unique circumstances of the pregnancy.}\DIFaddend\\
\bottomrule
\end{tabularx}
\caption{Examples of the differences between a randomly selected inappropriate argument and the improved versions generated by our models, sorted by the semantic similarity between the original and the rewrite.}
    \label{table-examples3}
\end{table*}
\clearpage

\textbf{Issue:} Ban plastic water bottles

\begin{table*}[h!]
\small
\centering
\begin{tabularx}{\textwidth}{sb}
    \toprule
\textbf{Model} & \textbf{Argument Improvement Differences} \\
\midrule
\textbf{LLaMA + Instruct.} + PPO$_{app<sim}$ & \vphantom{} We will be able to ban water bottles \DIFdelbegin \DIFdel{until }\DIFdelend \DIFaddbegin \DIFadd{once }\DIFaddend we get out of this recession\DIFdelbegin \DIFdel{!}\DIFdelend \DIFaddbegin \DIFadd{.}\DIFaddend \\\midrule
\textbf{LLaMA + Instruct.}                   & \vphantom{} We will be able to \DIFdelbegin \DIFdel{ban water bottles until we get out of this recession!}\DIFdelend \DIFaddbegin \DIFadd{phase out water bottles once we are no longer in this recession.}\DIFaddend \\\midrule
\textbf{LLaMA + Instruct.} + PPO$_{app=sim}$ & \vphantom{} We \DIFdelbegin \DIFdel{will }\DIFdelend \DIFaddbegin \DIFadd{won't }\DIFaddend be able to ban water bottles until we\DIFdelbegin \DIFdel{get out of this recession !}\DIFdelend \DIFaddbegin \DIFadd{'re able to generate enough income to replace them, especially if we're in a recession where we're not able to acquire funding for supplies.}\DIFaddend \\\midrule
\textbf{LLaMA + Instruct.} + PPO$_{app>sim}$ & \vphantom{} \DIFdelbegin \DIFdel{We will be able to ban water bottles until we get out of this recession!}\DIFdelend \DIFaddbegin \DIFadd{Given the current economic climate, prohibiting the use of single-use plastics such as water bottles may not be feasible until we are able to gain more economic stability.}\DIFaddend\\\midrule
\textbf{LLaMA + Instruct.} + PPO$_{app}$     & \vphantom{} We \DIFdelbegin \DIFdel{will be able to ban water bottles until we get out of this recession!}\DIFdelend \DIFaddbegin \DIFadd{may want to consider banning water bottles in order to reduce our plastic waste and dependency on single-use plastics when we can, although it may be difficult to do so until we are able to get back on our feet financially.}\DIFaddend \\
\bottomrule
\end{tabularx}
\caption{Examples of the differences between a randomly selected inappropriate argument and the improved versions generated by our models, sorted by the semantic similarity between the original and the rewrite.}
    \label{table-examples4}
\end{table*}

\textbf{Issue:} Tv is better than books

\begin{table*}[h!]
\small
\centering
\begin{tabularx}{\textwidth}{sb}
    \toprule
\textbf{Model} & \textbf{Argument Improvement Differences} \\
\midrule
\textbf{LLaMA + Instruct.} + PPO$_{app<sim}$ & \vphantom{} \DIFdelbegin \DIFdel{tv because only tv can bring u live news at books u can't find also it's educational }\DIFdelend \DIFaddbegin \DIFadd{Through television, one can access live news that books cannot offer, as well as educational programming.}\DIFaddend \\\midrule
\textbf{LLaMA + Instruct.}                   & \vphantom{} \DIFdelbegin \DIFdel{tv because only tv can bring u live news at books u can't find also it 's educational}\DIFdelend \DIFaddbegin \DIFadd{Television is a great way to stay informed of current events since it brings you live news coverage. Books, on the other hand, are not as accessible when it comes to quickly accessing up-to-date information}\DIFaddend \\\midrule
\textbf{LLaMA + Instruct.} + PPO$_{app=sim}$ & \vphantom{} tv \DIFdelbegin \DIFdel{because only tv can bring u live newsat books u can 't find also it's educational}\DIFdelend \DIFaddbegin \DIFadd{is a great source of live news, which is useful in situations where newspapers and websites cannot provide up-to-date information in a timely manner. Additionally, television can bring you news from around the world that you cannot find in books, which is also educational.}\DIFaddend \\\midrule
\textbf{LLaMA + Instruct.} + PPO$_{app>sim}$ & \vphantom{} \DIFdelbegin \DIFdel{tv because only tv can bring u live news at books u can 't find also it's educational}\DIFdelend \DIFaddbegin \DIFadd{With a computer, you can access news sites for up-to-date information, as well as gain access to a wide range of educational materials. Additionally, computers are more versatile than TVs, which are limited to broadcast news and can be more difficult to find certain information.}\DIFaddend \\\midrule
\textbf{LLaMA + Instruct.} + PPO$_{app}$     & \vphantom{} \DIFdelbegin \DIFdel{tv because only tv can bring u live news at books u can't find also it 's educational }\DIFdelend \DIFaddbegin \DIFadd{Television can be a great source of news and information as it allows you to access live coverage of events as they happen. Additionally, there are many educational programs available on television which can expand your knowledge and understanding.}\DIFaddend \\
\bottomrule
\end{tabularx}
\caption{Examples of the differences between a randomly selected inappropriate argument and the improved versions generated by our models, sorted by the semantic similarity between the original and the rewrite.}
    \label{table-examples5}
\end{table*}